\documentclass{article}
\usepackage[preprint,nonatbib]{nips_2018}
\usepackage[square, comma, numbers, sort]{natbib}
\usepackage[utf8]{inputenc} % allow utf-8 input
\usepackage[T1]{fontenc}    % use 8-bit T1 fonts
\usepackage{hyperref}       % hyperlinks
\usepackage{url}            % simple URL typesetting
\usepackage{booktabs}       % professional-quality tables
\usepackage{amsfonts}       % blackboard math symbols
\usepackage{nicefrac}       % compact symbols for 1/2, etc.
\usepackage{microtype}      % microtypography
\usepackage{times,graphicx} 
\usepackage{comment}

\title{PANDA: Facilitating Usable AI Development}
\author{
    \textbf{Jinyang Gao$^\dagger$, Wei Wang$^\dagger$, Meihui Zhang$^\diamond$, Gang Chen$^\ddagger$, H.V. Jagadish$^\ast$,} \\ 
    \textbf{Guoliang Li$^\oplus$, Teck Khim Ng$^\dagger$, Beng Chin Ooi$^\dagger$, Sheng Wang$^\dagger$, Jingren Zhou$^\star$}\\
    \\
	 $^\dagger$National University of Singapore \hspace{2mm}
	 $^\diamond$Beijing Institute of Technology \hspace{2mm} $^\ddagger$Zhejiang University\\
	 $^\ast$ University of Michigan
	 \hspace{2mm} $^\oplus$ Tsinghua University \hspace{2mm} 
	 $^\otimes$ Alibaba Group
	 \vspace{2mm} \\
	 $^\dagger$\{jinyang.gao, wangwei, ngtk, ooibc, wangsh\}@comp.nus.edu.sg,
	 $^\diamond$meihui\_zhang@bit.edu.cn \\  $^\ddagger$ cg@zju.edu.cn, $^\ast$jag@umich.edu, $^\oplus$liguoliang@tsinghua.edu.cn,  $^\otimes$jingren.zhou@alibaba-inc.com
}

\begin{document}
\maketitle

\begin{abstract}
Recent advances in artificial intelligence (AI) and machine learning have created a general perception that AI could be used to solve complex problems, and in some situations over-hyped as a tool that can be so easily used.
Unfortunately, the barrier to realization of mass adoption of AI on various business domains is too high because most domain experts have no background in AI. 
Developing AI applications involves multiple phases, namely data preparation, application modeling, and product deployment. 
The effort of AI research has been spent mostly on new AI models (in the model training stage) to improve the performance of benchmark tasks such as image recognition. 
Many other factors such as usability, efficiency and security of AI have not been well addressed, and therefore form a barrier to democratizing AI. 
Further, for many real world applications such as healthcare and autonomous driving, learning via huge amounts of possibility exploration is not feasible since humans are involved.  
In many complex applications such as healthcare, subject matter experts (e.g. Clinicians) are the ones who appreciate the importance of features that affect health, and their knowledge together with existing knowledge bases are critical to the end results.
In this paper, we take a new perspective on developing AI solutions, and present a solution for making AI usable.  We hope that this resolution will enable all subject matter experts (eg. Clinicians) to exploit AI like data scientists.

\end{abstract}

\section{Introduction}\label{sec:intro}

Recent advances in artificial intelligence (AI) and machine learning provide many opportunities in improving various applications, business practices and models. 
For example, AI-based solutions driven by Big Data have achieved human-level performance in computer vision and speech processing benchmarks.
The availability of data has caused the rapid development of new models, whose success further fuels the interest for exploiting data in decision making.  
It is therefore not surprising that we see an increasing desire to exploit of AI in application areas such as finance and healthcare.

However, there is a significant barrier realizing the mass adoption of AI applications. 
Developing an AI application involves multiple phases, namely data preparation, application modeling, and product deployment. 
In fact, the effort of AI researchers was spent mostly on new AI models to improve the performance of benchmark tasks, e.g. the ImageNet competition\cite{DBLP:journals/corr/RussakovskyDSKSMHKKBBF14}. 
Many other factors such as usability, efficiency and security of AI have not been well addressed, and therefore form a barrier to democratizing AI. 

\begin{figure}[htb]
\vspace{-6mm}
	\centering
	\includegraphics[width=0.9\textwidth]{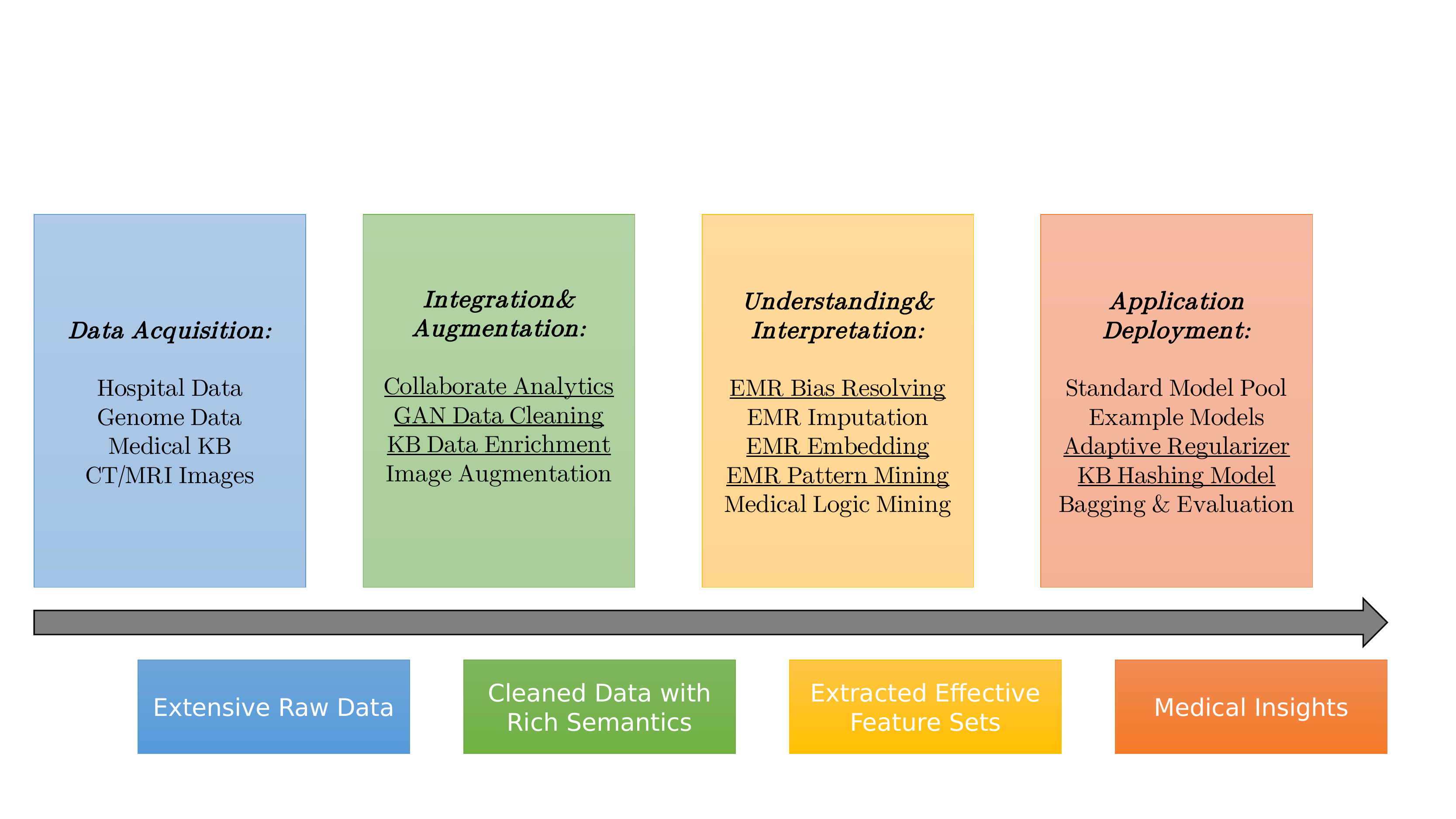}
	\caption{AI Development pipeline for Healthcare. \label{fig:hcflow}}
\end{figure}

We have been involved in developing the basic research, understanding and interpretation of requirements, to deployment and validation of several such applications. 
One example is the healthcare EMR (electronic medical record) application, where we worked with the clinicians, developed the model~\cite{cikm17}, validated the model, and integrated the application onto the production system after validation.
Figure~\ref{fig:hcflow} shows the development pipeline of a healthcare AI systems such as disease progression modelling.
Compared to devising a new algorithm on a standard benchmark problem, we face the following challenges:
1) There is no standard dataset for any application based on EMR. 
Data are biased, irregular and too noisy to be directly used as input for any ready-made model.
Exploration for new features could be very helpful for the specified application.
2) The usage of model and parameter setting strongly depends on a detailed application.
Finding a suitable solution requires both strong domain knowledge and machine learning background.
3) High stakes applications require strong reliability for the deployed product.

Based on our experience and observations, we examine the life-cycle of an AI application to locate the specific research topics related to the barrier from the perspective of AI and Big Data researchers, and application developers and data scientists. 

\begin{figure}[htb]
	\centering
	\includegraphics[width=0.25\textwidth]{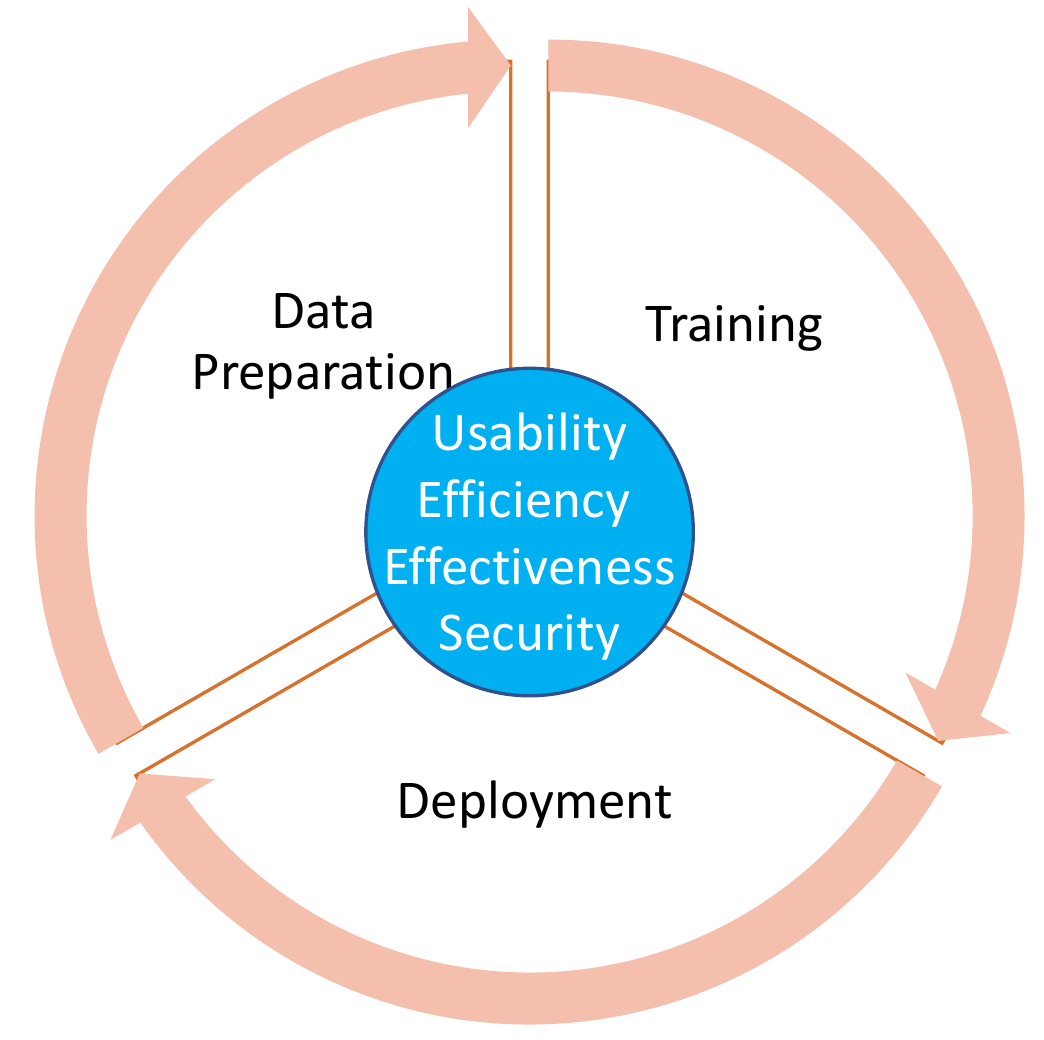}
	\caption{Development life-cycle of an AI application. \label{fig:life}}
\end{figure}

Succinctly, the development process of an AI application consists of three main stages (Figure~\ref{fig:life}): 
1) data preparation including acquisition, cleaning, labeling, integration and analytics; 
2) specified model design for the given application; 
3) product deployment which provides reliable service efficiently. 
Like the development of other applications, e.g. database applications, we need support from algorithms, models, tools and systems to ease the processing, reduce the cost, improve the performance and ensure the security of each stage of the development. 
In the remainder of this paper, we shall discuss each stage to analyze the challenges of realizing these goals while keeping in mind our goals: ease of use, effectiveness, efficiency, scalability and security.
%% AI, efficient AI, effective AI and secure AI.

\section{Data Preparation}

Industrial AI applications are often based on simple yet effective standard learning models. 
However, the development procedure is not trivial and often painful because of the quantity and quality of data. 
Unlike well-studied benchmark problems which come with a pre-defined training set, the training data of a real AI application is often not well pre-defined, cleaned or integrated. 
Data is so important for AI applications that it is referred to as the new oil. 
Currently, most datasets are created manually by domain experts or via crowdsourcing. 
The cleaning, integration, labeling and analyzing procedures are tedious and expensive for a large dataset. 

\subsection{Visualization and Interaction}

Tools with good usability can improve both efficiency and effectiveness. 
However, an easy-to-use system requires a lot of engineering work on the interface and functionalities.
%%% and research study on processing large scale data. 

We highlight the role of data visualization. 
A well-designed data visualization tool greatly assists the domain experts in reviewing the data, exploring the existing large-scale data, performing collaborative annotations, and effectively offering their expertise. 
Some new types of data visualization tools have been developed for this purpose, e.g. the interactive visualization tool with the collaborative annotation and recommendation functionalities. 
In addition, some crowdsourcing platforms allow the embedded Hypertext Markup Language to visualize the data in the micro-tasks. 
Other related research work, including the approximate visualization, the auto-ranked visualization, collaborative exploration, resolution reduction, explore by example, result recommendations, are extensively studied to make the input of domain experts’ knowledge more friendly. 
Efficiency of the visualization tools is also worthy of research, especially for big datasets, as it greatly affects the user experience.

Intelligent questioning modules must be designed to ask the right questions to extract useful information from domain experts who may not be conversant in AI techniques.  
Such kind of modules will not only serve to bridge the knowledge gap between domain experts and AI practitioners, they will also serve as gate keepers to ensure the validity of the data.  
The statistical distribution and bias of the data will have significant impact on the success of AI systems as most of them are data-driven.

\subsection{Cost-sensitive Acquisition}

We factorize the efficiency of data preparation into two parts following the efficiency definition of a training system~\cite{DBLP:journals/corr/HadjisZMR16}: hardware efficiency for the speed of processing one single (or batch) sample; statistical efficiency related to the total number of samples to process. 
The effectiveness refers to the quality of the pre-processed data and the performance of the model trained using the data.

\begin{figure}
	\centering
	\includegraphics[width=0.65\textwidth]{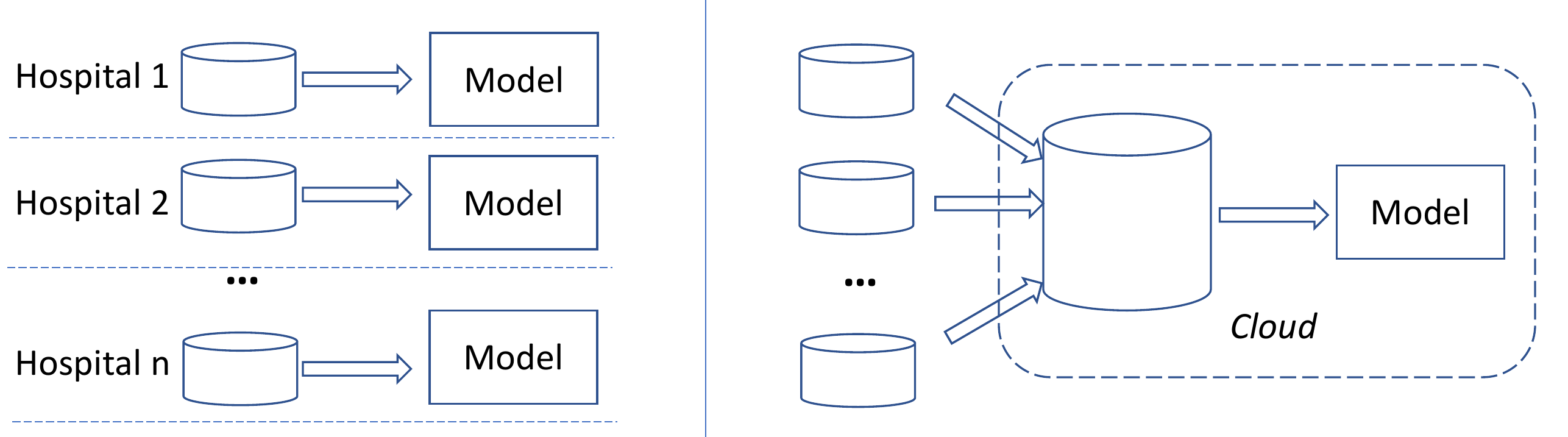}
	\caption{Collaborative analysis by integrating data from multiple hospitals.\label{fig:data}}
\end{figure}

To reduce the time spent for each sample, only those that require domain knowledge and can reduce the uncertainty of learned model should be presented to the domain experts. 
The interaction with domain experts may entail in all the processes in the pipeline where there exist uncertainties. 
The choice of step where interaction should be invoked and how much effort should be spent in it have to be optimized in a quantitative manner.

To reduce the total number of samples to process, we need to maximize the quantity of information instilled per sample. Then we can stop the processing early once we have enough knowledge about the data. The accuracy of the returned result is a critical issue. One way to increase the accuracy is to collect answers from different domain experts. However, in this case, some algorithms (e.g., the majority voting) have to be designed to reconcile the answers. 

\subsection{Data Privacy}\label{sec:data_security}

With the rapidly growing complexity of AI tasks, AI systems require more extensive cooperation among data providers and end-users. 
Data from multiple sources are required to be integrated and managed as a data ecosystem. 
We need a storage system to support collaborative analytics where different organizations having similar applications could share the common data processing flow while maintaining the confidentiality of the data. 
For example, to train an accurate model for medical image analysis, e.g thoracic disease identification based on x-ray images~\cite{2017arXiv171106373L}, hospitals have to collaborate to construct a large labeled image dataset (Figure~\ref{fig:data}).
In such a scenario, part of the data and its processing flow are required to be shared while the security for some other data and their relevant application model should be strictly protected. 
We have developed a rich semantic data management and storage system call ForkBase~\cite{DBLP:journals/corr/Wang18} based on the principle of immutability, sharing and security. Immutability ensures the traceability of data provenance. Sharing and security properties can facilitate the development for collaborative analytics.

\section{Application Modelling}\label{sec:train}

There is plenty of research on model architectures and training algorithms. However, implementing those ideas requires expertise knowledge about AI. 
Moreover, model selection and training configuration are typically done by experts with years of experience. All these together create a big barrier for AI application developers. 
%In this section, we introduce some services to automate and optimize the training process.

\subsection{Model Selection}

There are three different levels of AI developers, namely, AI researchers, AI beginners and domain experts. To enable all developers to train models efficiently and effectively, research on programming abstraction, resource management and user-system interaction is necessary. 

\begin{figure}
	\centering
	\includegraphics[width=0.65\textwidth]{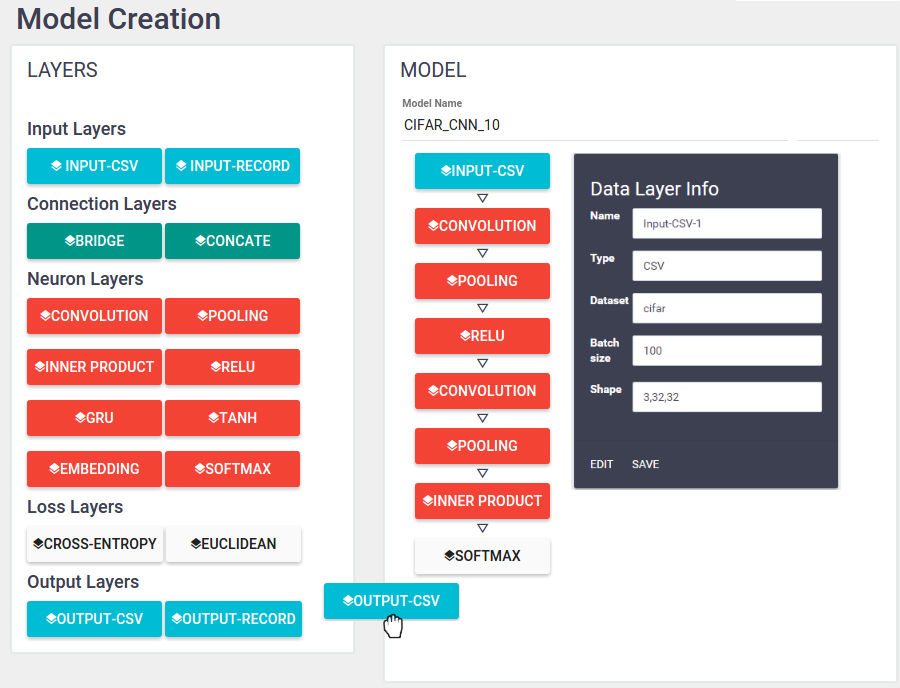}
	\caption{Drag-and-drop interface for model construction.\label{fig:drag}}
\end{figure}

For AI researchers, they are able to construct their own models using open-source libraries like Tensorflow~\cite{199317}. 
However, it is still tedious for them to tune many hyper-parameters of the training algorithms, including learning rate, total number of training iterations, etc. 
In addition, they have to manage many intermediate models and results. 
In fact, the checkpoint files for model parameters generated during the training are large for big models such as VGG~\cite{DBLP:journals/corr/SimonyanZ14a}. 
A tool with distributed hyper-parameter search, e.g. based on Bayesian optimization~\cite{NIPS2012_4522} or random search~\cite{Bergstra:2012:RSH:2188385.2188395}, is desired. 
A model management database with model compression would save a lot of space and time for developers.

For AI beginners, a simple, flexible, and extensible interface or programming abstraction is vital for them to get started.
Many open source libraries with good programming abstractions have been developed, including Keras\footnote{https://keras.io/} and scikit-learn\footnote{http://scikit-learn.org/}, which are widely used by students to learn data science and deep learning. 
A more convenient interface for beginners would be like drag-and-drop or plug-and-play on web pages as shown in Figure~\ref{fig:drag}, which sends the models back to the servers for training and tuning automatically. 

For domain experts, they know the data well. 
However, they may have little knowledge about the AI models and training algorithms. 
Therefore, it would be better to just let them prepare the data and specify the task. 
To implement such a system, we need to provide built-in models and model selection algorithms. 
In fact, many AI applications share the similar models. For example, convolutional neural networks (CNN)~\cite{DBLP:conf/nips/KrizhevskySH12} are the backbone models for image classification tasks, including vehicle classification, flower classification, food classification, etc. 
We can also implement other popular models (like LSTM~\cite{lstm}, CapsuleNet\cite{DBLP:journals/corr/abs-1710-09829}) as built-in models and share them for different applications. 
There are also multiple models for the same task. 
For example, InceptionNet~\cite{DBLP:journals/corr/SzegedyIV16}, ResNet~\cite{DBLP:journals/corr/HeZRS15} and SqueezeNet~\cite{DBLP:journals/corr/IandolaMAHDK16} are all CNN models for image classification. 
However, they have different characteristics, where some models are more accurate but more resource hungry. 
Model selection is a research problem~\cite{DBLP:journals/corr/abs-1708-07308}, which trades off between efficiency (i.e. speed and memory) and effectiveness (i.e. accuracy).

\begin{figure}
	\centering
	\includegraphics[width=0.3\textwidth]{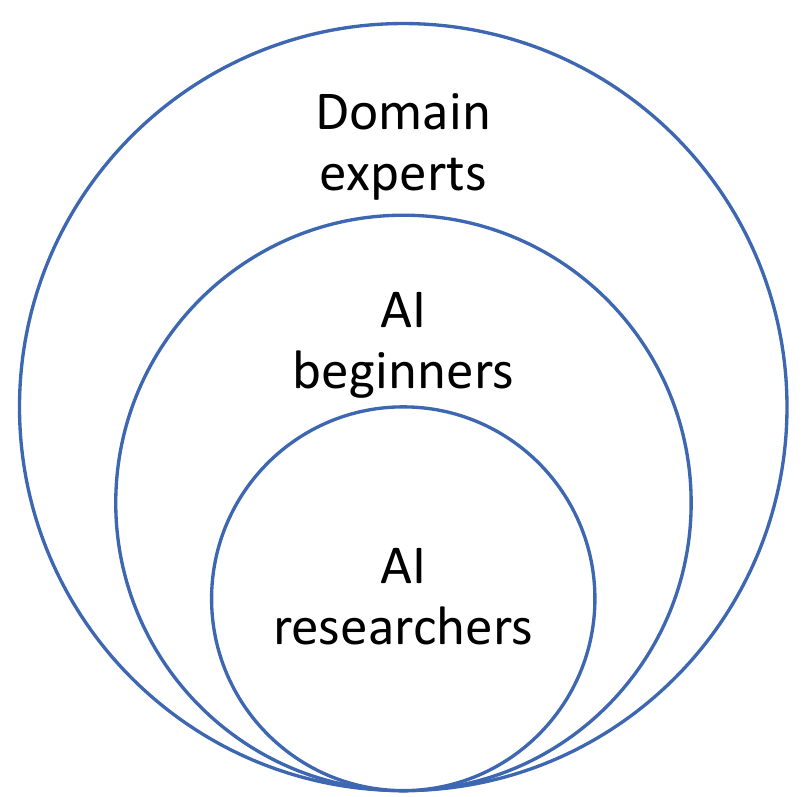}
	\caption{Optimization scope for different AI developers.\label{fig:level}}
\end{figure}

The features related to usability for the three types of AI developers are summarized in Figure~\ref{fig:level}. 
The features from the inner circles benefit developers in the same circle and in the outer circles.

\subsection{Cost-sensitive Modelling}

The recent resurgence of AI is mainly driven by deep learning, which expands traditional machine learning models with more complex structures to increase their capability of modeling data. 
From the statistics of a famous visual recognition challenge, ILSVRC, the number of layers of the annual winning model increases from 8 layers in 2012 to 152 layers in 2015.
For example, the number of layers of deep convolutional neural networks (CNN) have reached one thousand~\cite{DBLP:journals/corr/HeZRS15}. DeepForest~\cite{DBLP:journals/corr/ZhouF17} model stacks multiple random forests together. Models for text comprehension including question answering, typically combine many recurrent neural networks with attention modeling~\cite{Wang2017GatedSN}. New models, like CapsuleNet~\cite{DBLP:journals/corr/abs-1710-09829}, are also very complex in terms of the operations and number of parameters. 
At the same time, the training dataset size is also increasing sharply. On the one hand, big datasets are required by complex models to avoid over-fitting. On the other hand, big datasets need large models to capture the complex data regularities. Thereafter, datasets and models are affected by each other, and both grow in size and complexity.
We do see better performance (i.e accuracy) as a consequence. 
However, we also notice the efficiency cost in terms of computation, memory and disk cost. 
Model compression~\cite{DBLP:journals/corr/IandolaMAHDK16} replaces some complex structures in the model architecture with simple ones. 
For instance, fully connected layers in CNNs are replaced with fully convolutional layers\cite{DBLP:journals/corr/ShelhamerLD16}. Squared convolution filters are factorized into 1-dimensional convolution filters~\cite{DBLP:journals/corr/SzegedyIV16}. Bottleneck convolution layers are also widely used~\cite{DBLP:journals/corr/SzegedyIV16}. 
Architecture optimization (or search) for efficiency (without deteriorating the performance) is now mainly done based on experience and trial-and-error.

To train such a cost sensitive model selected by the above process, reducing the high demand on training data processing cost is also an active research problem.
There are three directions in general:
reducing the processing cost per visit of training sample; 
reducing the number of training samples; and reducing the number of visits per training sample (i.e. number of iterations);
Feature hashing and embedding methods~~\cite{ccfh,attembed} could be used to reduce the cost per training sample.
Few-shot learning~\cite{oneshot}, meta-learning and transfer learning~\cite{pansurvey} are major solutions towards reducing the number of samples need to be trained.
Adaptive and importance based sampling methods~\cite{activesample,variancesample} lead to faster convergence and less visits per training sample. In some extreme cases, samples which are not informative could be even removed from the training process without sacrificing model accuracy. 
%Research on model compression and convergence acceleration have achieved good progress. 
 
Genetic algorithm, reinforcement learning and model-based optimization~\cite{2017arXiv171200559L} may help make some progress. 
Researchers have been constructing more and more complex models in recent years. 
%%% It is time to re-examine the direction. 
However, simple models usually have the advantages of good interpretation. 
Hence, new models with simple structures and comparable performance, are also worth to investigate. 
Convergence acceleration that reduces the total number of training iterations is a difficult problem. 
Approaches for some special models have been proposed, e.g. increasing the batch-size\cite{DBLP:journals/corr/abs-1708-03888}. 
It remains challenging for (asynchronous) distributed training due to gradient staleness~\cite{2013arXiv1312.7869W} caused by communication delay. 

%%% ooibc: below is still not smooth
Trading off between efficiency and effectiveness has never been easy. 
In practice, what to optimize depends on the expectation of the users or the requirement of the applications. 
Notwithstanding, it is also related to fairness or resource saving. 
%%% ooibc: fairness of what
Typically, the improvement at the final stage of training is usually very minor, e.g. from accuracy 99\% to 99.2\%. 
When the hardware resource is shared by multiple tenants as investigated by \cite{DBLP:journals/corr/abs-1708-07308}, the cluster administrator can stop such instances to release the GPU to train other users' models. 
When the model is running on cloud platforms e.g. Amazon EC2, the running time is directly related to the fees. 
User expectation, application requirement, cost and the fairness are metrics to consider for the stopping criteria.

\subsection{Auto-tuning Models based on Knowledge-bases}

%Since most AI researchers are working towards more effective AI models by proposing or tuning the architecture, loss function, feature transformation, etc., we shall not discuss much on these aspects. 
%Here we discuss the idea of combining domain knowledge into the model construction.
Building domain specific knowledge base has been widely accepted as the foundation of conducting domain specific analytics. 
However, there is no golden standard as to what kind of knowledge base should be constructed and how they should be utilized to improve the analytic model. 
Currently, a knowledge base is mostly used in simple tasks such as manual analytics and visualization. 
There is no doubt that a domain specific knowledge base should be a valuable resource for all kinds of applications. 
However, it is still not clear how it can directly benefit applications based on complex models such as deep learning.
Intuitively, using domain specific knowledge base to improve a machine learning model is a paradox: knowledge base records how entities/features are related (usually in qualitative manner), while machine learning models tend to learn those relations from the training data (usually in quantitative manner). 
The main challenge of applying a knowledge base is how to balance the qualitative relations and quantitative relations.
We believe that using the qualitative relations from knowledge base as a prior distribution (i.e. regularization term) could be a simple, general, feasible solution. 
Nowadays, typical regularization methods are mostly acting in quantitative manner. 
%%There has been an attempt to 
For the healthcare system mentioned in Section 1, we designed a regularization term based on healthcare domain knowledge. 
However, the domain knowledge used there is limited to ontology knowledge, and the regularization method designed is limited to a certain kind of classification task.
%%% Therefore, 
Logically,
using knowledge for regularization requires research from two areas. 
One is to build a knowledge base that can clearly describe qualitative relations among features/samples. 
Another is to design regularization methods that can work in a qualitative manner.

\section{Model Deployment}\label{sec:deploy}

Models are trained offline and then deployed on cloud platforms, dedicated servers or edge devices for online predictions. Most research focuses on model training. 
In fact, the deployment process is not any simpler than training. It involves much engineering work, e.g. fault tolerance and load balance. These are also interesting research topics. 

%Here we shall discuss some research problems for making model deployment easier, and making the predictions faster, better and securer. 

\subsection{Reliability and Interpretability}

AI applications must go through a sequence of checks and validations before deployment. Once an application is deployed, we still need to monitor the performance, scale the throughput according to the demands, keep the load balance and recover nodes from failures. 
A one-step deployment service that combines and automates all these operations together is helpful. 
Besides automation, we highlight the importance of reliability and interpretability of models for the usability of model deployment.

Vertical domains like healthcare and finance have demanding requirements on the reliability of deployed applications. 
A simple solution is to monitor the performance and switch the working mode to human mode when AI is uncertain about some requests. 
Most machine learning models are soft margin based and have a self-evaluation for its accuracy (e.g. the Softmax outputs in logistics regression and most deep neural networks). 
However, this self-evaluation is only accurate when there is no concept drifting and the data characteristic exactly matches with the model assumptions. 
For example, for an application that is based on Naive Bayes model, when the input features appear are highly correlated, the real accuracy will drop significantly while its self-evaluation will have almost 100\% confidence about its prediction. Therefore, the self-evaluation may not be reliable. Designing a robust model to monitor the system performance is thus necessary. For example, we may continuously check the data distribution to see if the characteristic matches with the model assumptions. We can also collect feedback from users to evaluate the performance of the deployed model.

In addition to reliability, interpretation is also important.
For example,
doctors often ask the question---``how is the prediction generated?".
Explanations are essential for the democratization of AI on critical applications. 
Most complex machine learning models work like a black-box. 
Even for their designers, it is difficult to know the exact reason for every decision or prediction. Using black-box systems to do critical decisions or predictions could bring users a sense of distrust, violate regulation requirements and put the domain practitioners in a competitive relationship with AI solutions. 
All these factors are harmful for the success of AI on these valuable applications. Explanations could significantly reduce outside resistance and hence ease the usage of AI systems. 
In healthcare applications, researchers working on computational phenotyping~\cite{emrphenotyping,marble,enrbm} are trying to find out the explainable risk factors from the models for healthcare problems.
There is a trend of research on model interpretation~\cite{DBLP:journals/corr/BailisORZ17,Stoica:EECS-2017-159,2017arXiv170304730K,DBLP:journals/debu/0001ZO17}. 

We aim to design a set of general mechanisms to make AI solutions more understandable for model designers, domain experts, regulators and end-users. 
For model designers, the explanation could help them to refine the model architecture and training process. 
For domain experts, the explanation could bring more insights and hence enhance the cooperative relationship. 
For regulators, the explanations could help them solve legal issues and build accountability systems. 
For end-users, explanations increase the quality of service and promote trust.

%%% ooibc: check below again
Working towards this direction,
we use a neural network as a research prototype, evaluate the importance and meaning for each neuron and analyze how they interact. 
Without loss of generality, this evaluation framework can be extended to any machine learning models whose data transformation process can be described as a graph (e.g. PGM and topic modeling). 
We conduct the evaluation via a novel concept called neuron saliency, which measures neuron efficiency in neural networks. 
By estimating neuron saliency, we are able to find out whether the basic unit of neural networks, namely the neuron, is contributing to the success of these models or other neurons. 
We first unify the neural networks in neuron representation and introduce dropout optimization for neural networks. Then two methods are proposed to estimate neuron saliency efficiently by dropout and gradient information respectively. Based on the neuron saliency, algorithms for optimizing the training of neural networks are developed, and in the meantime, a novel algorithm for model compression by dropping low saliency neurons is introduced.

\subsection{Cost-sensitive Deployment}

%However, because of increased power consumption and memory usage, it is impractical to deploy such models on embedded devices with limited hardware resources and power constraints.reduce the model complexity without sacrificing accuracy

Efficiency or latency is more critical for the deployment stage than other stages as this stage is online. 
For example, because of the high requirement on latency (less than 1ms) during database querying, the learned index~\cite{2017arXiv171201208K} has to replace the inference code from a Tensorflow implementation with hand-crafted but well optimized code, even for a simple neural network model. 
Optimization should be conducted from every strata~\cite{DBLP:journals/corr/BailisORZ17,wangsigmodrec} including hardware, compiler, code, algorithm and models. GPU is excellent for training, but it costs extra time to transfer data from CPU to GPU. Hence, FPGA~\cite{DBLP:journals/corr/HanKMHLLXLYWYD16} has been applied as a replacement. 
To make it usable for non-FPGA programmers, we need a library with optimized operations (e.g. convolution for deep learning models) on FPGA, and a tool to convert the model trained on GPU to work on FPGA. Compilers like XLA\footnote{https://www.tensorflow.org/performance/xla/} and  Weld\cite{DBLP:journals/corr/abs-1709-06416} are designed to optimize AI and data analytic operations. Model compression that reduces the memory and computation cost for deploying models on small devices is a hot research topic~\cite{DBLP:journals/corr/abs-1710-09282}. The Tensor Train line of work is an example of such kind of model reduction effort~\cite{tensor_train}. 
The challenge is how to compress the model without sacrificing the accuracy. A more challenging but preferred solution is to design a new and simpler model directly to replace current big models. 
For example, it is desirable to replace CNNs with a new model with good interpretability, less computation and memory cost.
 
\begin{figure}
	\centering
	\includegraphics[width=0.6\textwidth]{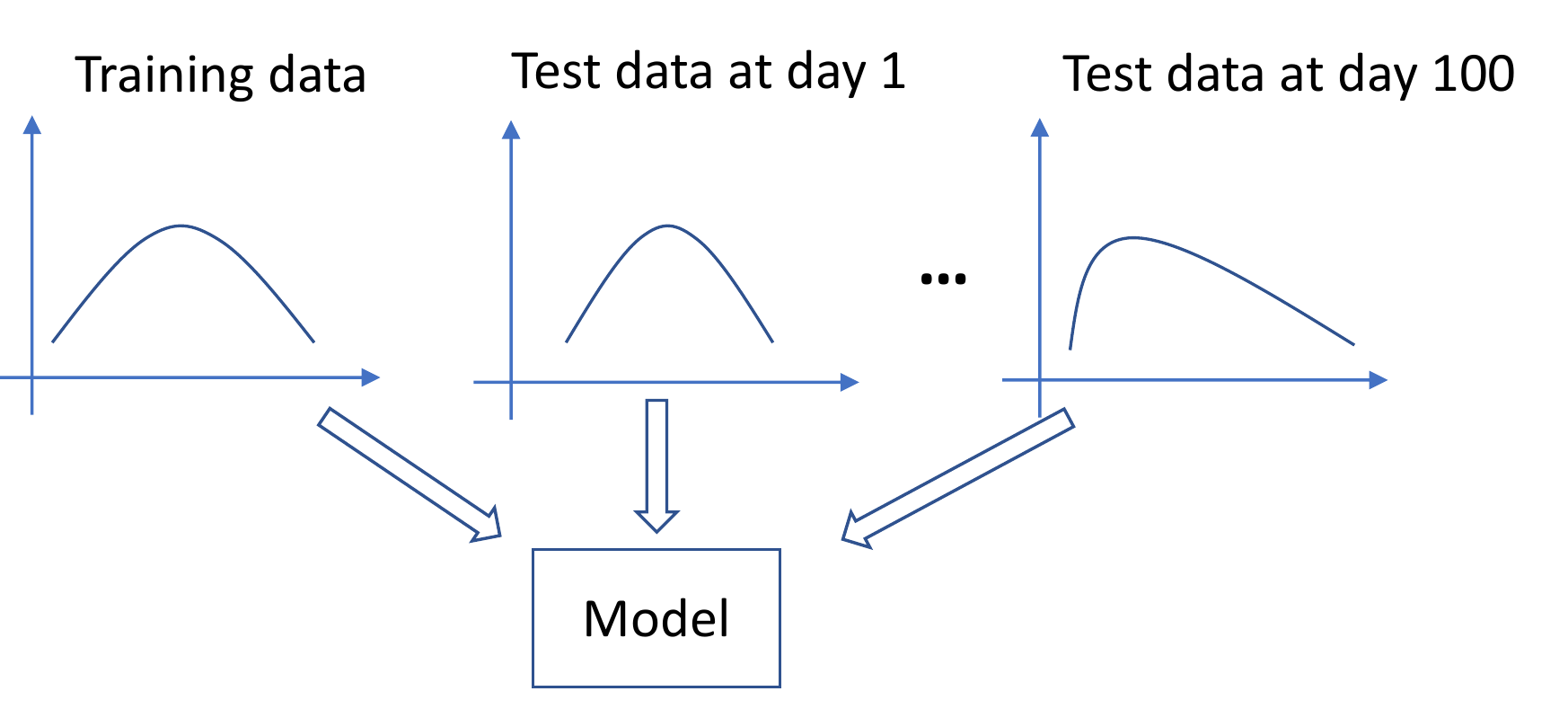}
	\caption{Illustration of data distribution drift.\label{fig:drift}}
\end{figure} 
 
% efficiency vs effectivenss of cnn models: https://arxiv.org/pdf/1605.07678.pdf
 
In terms of effectiveness, the most obvious issue is the change of data distribution as illustrated by Figure~\ref{fig:drift}. 
When the data distribution is evolving, the model should adjust to keep its performance. 

Continuous learning in nowadays deep learning context could be very challenging.
If a single model is used for the prediction, the only choice here is to design a transfer learning or online learning model that can leverage the online data to refine itself. 
It is still not clear how a model training using stochastic gradient descent can be efficiently updated or trained with stream data with performance guarantee.
An alternative solution that is commonly adopted in practice is ensemble modeling. 
Instead of using a single prediction model that may suffer from over-fitting and change of data distribution, ensemble modeling is a more robust solution since the final prediction are based on the output of several different models. 
However, simply averaging the results of multiple models can only result in a static robust model (i.e. less sensitive to over-fitting) but it still cannot adapt to the change of data distribution such as concept drift. 
In many cases the best model that should be trusted depends on the data distribution of the online incoming data. To get the best performance from multiple models, inference based on real-time feedback is strongly required. For both solutions, we have to optimize the cost in terms of power consumption and storage as the target devices may be mobile phones or IoTs.

\subsection{Security}

Nowadays, many applications are deployed on cloud platforms, e.g. Amazon EC2. Users submit their request to the cloud platforms for processing and then receive the prediction results. For such cases, we need to protect both the request (or query) data and the model to avoid leaking training data. To protect the request data, we have to encrypt it. 
Therefore, the models must accept encrypted data as input and generate encrypted predictions. 
Similar to the approaches for training over encrypted data, inference~\cite{DBLP:journals/corr/XieBFGLN14,DBLP:journals/corr/abs-1711-05189} over encrypted data is mainly based on homomorphic encryption. 
Considering that the efficiency problem is more critical for inference than training, approaches with fast inference speed is necessary.
To protect the model, we typically add noise to prevent users from inferring some properties of the training data. For example, users can infer the membership of a certain data sample based on the prediction accuracy and confidence. In particular, if the model is over-fitting on the training data and is very confident about a test data sample, it is likely that this sample is included in the training dataset. However, adding noise into the prediction results would affect the accuracy of the model from the user's perspective. 
A research direction is to train a model with good generalization ability such that it performs equally well on both training and testing data. Then, we cannot infer the membership of the test data. In fact, it is a shared research goal from the perspective of security and machine learning training.

\section{PANDA Solution}

We have been developing systems towards resolving the issues discussed in this paper. 
We shall now discuss the PANDA architecture that we believe can address the issues highlighted in this paper.

\subsection{Basic End-to-end Analytics Stack}

\begin{figure}[t]
	\centering
	\includegraphics[width=0.8\textwidth]{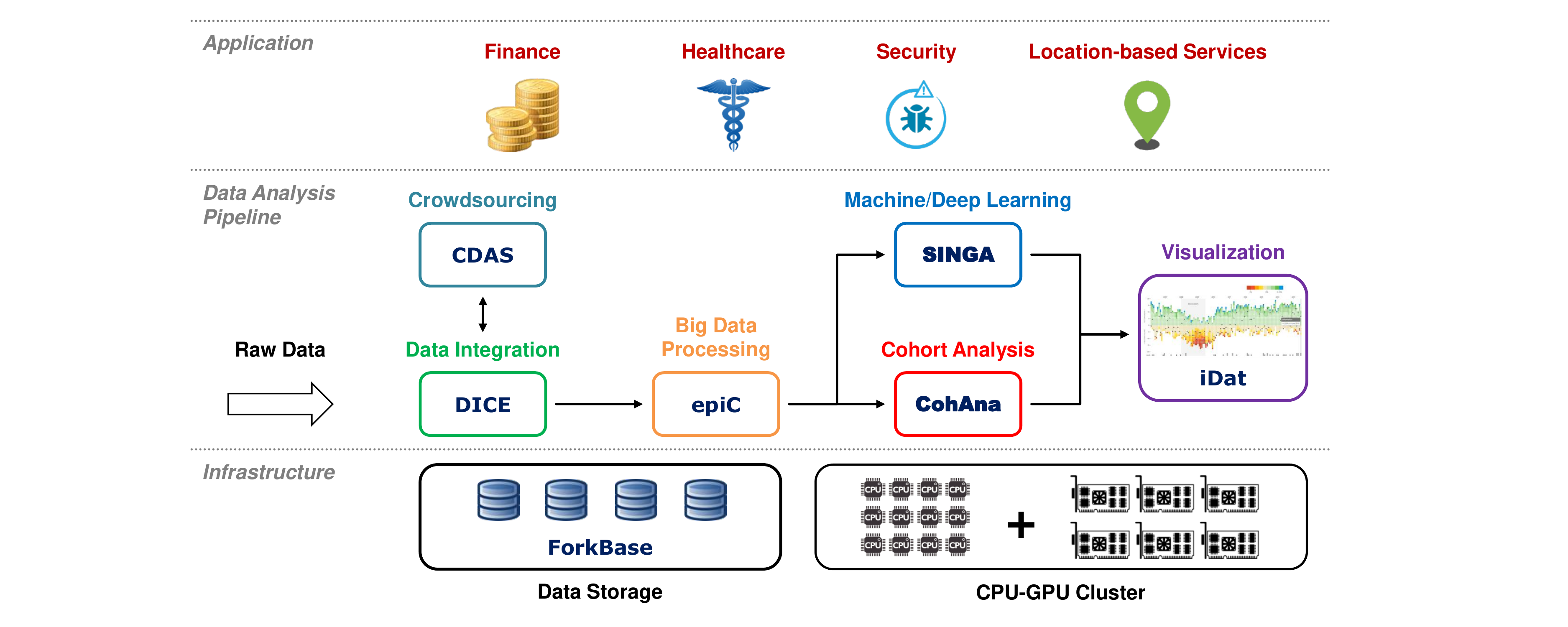}
	\caption{An end-to-end analytics system stack.\label{fig:stack}}
\end{figure}

Figure~\ref{fig:stack} shows the current stack of our systems. Healthcare is one of our primary applications. We are collaborating with multiple local hospitals, who give us the data and help to validate our results. CDAS~\cite{DBLP:journals/corr/abs-1207-0143} is a crowdsourcing system used by doctors to add their knowledge into the data, e.g. by labeling. DICE is a system for data integration that cleans raw EMR data based on expert defined rules. epiC~\cite{epic} is our large scale batch processing engine. ForkBase\footnote{ForkBase is the second version of UStore~\cite{DBLP:journals/corr/DinhWWCCLORTXZZ17}, which has evolved substantially since the first implementation}~\cite{DBLP:journals/corr/Wang18} is our data storage engine designed with rich semantics and three key properties, namely immutability, sharing and security. 
After labeling, cleaning and pre-processing, the data is fed into training or analytics engines. Apache SINGA~\cite{mm15} is a deep learning platform initiated by us, which focuses on memory and speed efficiency optimization. On top of SINGA, we have a platform called Rafiki~\cite{2018arXiv180406087W}, which provides training and deployment services for analytics tasks. CohAna~\cite{DBLP:journals/corr/JiangCCOTT16} is a cohort analysis engine designed for tasks like customer churn analysis. 
iDat is a visualization tool for presenting the analytics results and users to explore the results. 
Other applications include finance data analytics and cyber-security.

\subsection{Specific Challenges as Plug-ins}

In the PANDA system, we aim to ease the development of AI applications when facing the aforementioned challenges. However, not all of them are shared by every application. Different applications could benefit from different workflows. Therefore, instead of building the solution for each challenge as a fixed step in the pipeline, we propose to build them as optional plug-ins.

Each plug-in consists of three basic components.
We always build one simple default solution which is general and data insensitive.
To resolve the specified challenges, a detector is developed to examine whether the inputs follow certain data characteristics. 
If the answer is true, a specific solution, which is typically data-driven, will be applied to replace the default solution.

We use the knowledge-based regularization module in the pipeline as an illustrative example.
There will be a detector to examine whether the parameter set are correlated to a concept set. Once such connection is verified, the regularization term will be constructed based on the relations of concepts in the knowledge-base.
Meanwhile, the default solution is just a simple L2-norm, which will be applied to most parameter sets that are without any meaningful relation to existing concepts. 

\subsection{Key Modules}

\subsubsection{Application Driven Data Exploration}

We propose to build an automatic feature and sample exploration model. 
Given a budget and a pricing model for the dataset, the target of this model is to find an iterative and explorative data acquisition strategy to obtain the subset of data which has a price lower than the budget and leads to the best application performance. 
This problem can be viewed as a generalized process of active learning, which only optimizes the application performance by selecting data samples. 
However, if the data acquisition model can decide the set of features they should query for data samples, there are more opportunities to further reduce the data acquisition cost than simply applying active learning. 
This process is also different from feature engineering which aims to filter out noisy features, since in our exploration model the filter is not only cost sensitive, but also is designed to filter out as many unnecessary features as possible.
%%% ooibc2: pls check the above

Data quality management is also necessary, but acts as a fundamental module to support the data exploration process. Data exploration can be viewed as a set of cost-sensitive schemes for data acquisition, where the data quality is managed in a proactive manner. Typical data quality management consists of evaluations of data quality (or system performance) and optimizations (e.g. data cleaning) for data quality based on the given dataset. 
However, its achievement is typically limited if there exist fundamental imperfections in the given dataset. 
Instead of passively recovering noisy features or labels and inferring them based on uninformative samples, data exploration looks for indicative features and learns only from representative and valuable samples. 

\subsubsection{Data Driven Model Selection}

Data quality affects the model performance directly. 
However, although the data is well pre-processed, the model may still perform poorly. 
This is because, when applying a machine learning model to a dataset to solve an application, we are actually making a set of assumptions on the characteristics of the dataset. 
For example, using Naive Bayes model means we are assuming that the features of a data sample should be independent when given the labels, and using logistics regression means we are assuming that the labels are generated based on a linear combination of all its features.
In terms of regularization, using L2-norm regularization means we are assuming a Gaussian prior over the model parameters learned through the datasets.
If there is a mismatch between the real characteristics of the dataset and the model assumptions, the performance suffers.
This is the critical reason why we design or select different machine learning models for different applications.
Based on the above intuition, avoiding the mismatch between data characteristics and model assumption could be an efficient way to automate model designing and tuning. 

For each model or processing step, we propose to build a matching evaluation to test if the input data follow the model assumption. 
If there is a mismatch, data transformations need to be applied. 
For example, if the model assumes the data should be linearly separable (e.g. SVM), but the data are not linearly separable (this evaluation can be easily obtained from the optimization result of SVM), two choices should be recommended instead of directly applying SVM: using feature engineering to pre-process the data to make it linearly separable; or applying kernel SVM which does not assume that the data are linearly separable.
The examination of the matching between data characteristics and model assumption could significantly ease the model designing and tuning process or even automate it, since making the exact match is exactly one of the principles of model designing. 

\subsubsection{Reliable Answers for High Stake Applications}

For most of the complex data analytics and decision making problems such as financial investment~\cite{app_stock}, medical treatment~\cite{survey_medical} and self-driving system~\cite{app_driving}, learning algorithms, while in progressive development, are widely believed to be able to surpass human performance in the near future.
Such emerging ``high stakes'' applications of AI pose exacting demands on the reliability of deployed solutions.
However, most of these applications rely on prevailing deep neural network models~\cite{nature_dl}.
While these models may provide high prediction accuracy in the general case, they may be vulnerable to unexpected egregious errors~\cite{exp_fool,model_fool,exp_noise}, particularly when being applied to data points that are not well-represented in the training set.
In some cases, the deep learning models are no better than random guesses on regions lacking of training points, and yet predict with high confidence.
For high stakes applications, every decision matters and such irresponsible actions are definitely prohibited.
Unfortunately, most deep learning models act like a black-box without much explanation, and are hard to understand even for domain experts~\cite{exp_zfnet}.
It is not practical to prevent such failures by manually examining the logic inside a deep learning solution.
Consequently,
developing a deep learning solution with reliable behaviour has attracted a great deal of interest.

Instead of answering all the problems, we propose that a deep learning model should only be responsible to answer problems which it has been trained to answer, i.e. problems that lie in the reliable region. 
The reliable region is defined as a data distribution generalized from the training set where the deep learning model can achieve as good  performance as when tested on the training set.
Reliable deep learning solutions are useful for many high stakes applications.
For example, a model for CT image classification with 90\% accuracy cannot be applied onto any other clinical system, where the typical minimum requirement is 95\%. 
As an alternative, by applying a model that provides 99\% accuracy inside a reliable region which covers half of the patient images, the workload of radiologists could be effectively reduced by half.
Developing an all-weather strategy for financial investment to significantly outperform the market is usually not practical. 
However, if we can build a model which significantly outperforms the market within a small reliable region, safe arbitrage can be done when the opportunity arises (i.e. market state falls within a reliable region).

%\section{AI as a Service}
%We acknowledge that the challenges described in this paper require research from various domains. We are working towards to resolve some issues by developing a AI as a Service (AIaaS) platform on top of infrastructure as a service (IaaS), e.g. Amazon AWS.  By putting AIaaS on cloud platforms, developers are exempted from hardware resource management. They would also enjoy the features like elasticity and scalablility. As shown in Figure~\ref{fig:arch}, we extend the architecture of Clipper~\cite{201468} by adding data service and training service. Each stack of services deals with the issues from one stage of the development life-cycle. With these services, we aim to reduce the general efforts and technical debts~\cite{43146} of developing domain specific AI solutions.

\section{Conclusions}

There is no doubt that AI technologies will have great success in many vertical domains in the next few years. 
However, the mass production of AI poses many challenges for the current data analytics pipeline and other support system infrastructure, especially for critical decision making in a domain specific problem. 
In this paper, we review some challenges with respect to the issue of usability, efficiency and effectiveness and security in data preparation, training and product delivery phases of an AI application. 
Compared to the great success achieved in recent benchmark problem (e.g. CV and NLP) modeling, these challenges are not well addressed by current AI research but play a vital role in practical domain specific AI solution development. 
We summarize several research directions and discuss some preliminary methods. 
We are developing an AI platform called PANDA to resolve the aforementioned issues and support fast development of domain specific applications. We hope to make AI more usable, explainable, and scalable.


\begin{thebibliography}{10}

\bibitem{199317}
M.~Abadi, P.~Barham, J.~Chen, Z.~Chen, A.~Davis, J.~Dean, M.~Devin,
  S.~Ghemawat, G.~Irving, M.~Isard, M.~Kudlur, J.~Levenberg, R.~Monga,
  S.~Moore, D.~G. Murray, B.~Steiner, P.~Tucker, V.~Vasudevan, P.~Warden,
  M.~Wicke, Y.~Yu, and X.~Zheng.
\newblock Tensorflow: A system for large-scale machine learning.
\newblock In {\em OSDI}, 2016.

\bibitem{variancesample}
G.~Alain, A.~Lamb, C.~Sankar, A.~Courville, and Y.~Bengio.
\newblock Variance reduction in sgd by distributed importance sampling.
\newblock {\em arXiv preprint arXiv:1511.06481}, 2015.

\bibitem{DBLP:journals/corr/BailisORZ17}
P.~Bailis, K.~Olukoton, C.~R{\'e}, and M.~Zaharia.
\newblock Infrastructure for usable machine learning: The stanford dawn
  project.
\newblock {\em arXiv preprint arXiv:1705.07538}, 2017.

\bibitem{Bergstra:2012:RSH:2188385.2188395}
J.~Bergstra and Y.~Bengio.
\newblock Random search for hyper-parameter optimization.
\newblock {\em JMLR}, 2012.

\bibitem{attembed}
X.~Cai, J.~Gao, K.~Y. Ngiam, B.~C. Ooi, Y.~Zhang, and X.~Yuan.
\newblock Medical concept embedding with time-aware attention.
\newblock In {\em IJCAI}, 2018.

\bibitem{app_driving}
C.~Chen, A.~Seff, A.~Kornhauser, and J.~Xiao.
\newblock Deepdriving: Learning affordance for direct perception in autonomous
  driving.
\newblock In {\em ICCV}, 2015.

\bibitem{DBLP:journals/corr/abs-1710-09282}
Y.~Cheng, D.~Wang, P.~Zhou, and T.~Zhang.
\newblock A survey of model compression and acceleration for deep neural
  networks.
\newblock {\em arXiv preprint arXiv:1710.09282}, 2017.

\bibitem{app_stock}
X.~Ding, Y.~Zhang, T.~Liu, and J.~Duan.
\newblock Deep learning for event-driven stock prediction.
\newblock In {\em IJCAI}, 2015.

\bibitem{DBLP:journals/corr/DinhWWCCLORTXZZ17}
A.~Dinh, J.~Wang, S.~Wang, G.~Chen, W.-N. Chin, Q.~Lin, B.~C. Ooi, P.~Ruan,
  K.-L. Tan, Z.~Xie, et~al.
\newblock Ustore: a distributed storage with rich semantics.
\newblock {\em arXiv preprint arXiv:1702.02799}, 2017.

\bibitem{exp_noise}
A.~Fawzi, D.~Moosavi, M.~Seyed, and P.~Frossard.
\newblock Robustness of classifiers: from adversarial to random noise.
\newblock In {\em NIPS}, 2016.

\bibitem{oneshot}
L.~Fei-Fei, R.~Fergus, and P.~Perona.
\newblock One-shot learning of object categories.
\newblock {\em TPAMI}, 2006.

\bibitem{activesample}
J.~Gao, H.~Jagadish, and B.~C. Ooi.
\newblock Active sampler: Light-weight accelerator for complex data analytics
  at scale.
\newblock {\em arXiv preprint arXiv:1512.03880}, 2015.

\bibitem{ccfh}
J.~Gao, W.-c. Lee, Y.~Shen, and B.~C. Ooi.
\newblock Cuckoo feature hashing: Dynamic weight sharing for sparse analytics.
\newblock In {\em IJCAI}, 2018.

\bibitem{DBLP:journals/corr/HadjisZMR16}
S.~Hadjis, C.~Zhang, I.~Mitliagkas, D.~Iter, and C.~R{\'e}.
\newblock Omnivore: An optimizer for multi-device deep learning on cpus and
  gpus.
\newblock {\em arXiv preprint arXiv:1606.04487}, 2016.

\bibitem{DBLP:journals/corr/HanKMHLLXLYWYD16}
S.~Han, J.~Kang, H.~Mao, Y.~Hu, X.~Li, Y.~Li, D.~Xie, H.~Luo, S.~Yao, Y.~Wang,
  et~al.
\newblock Ese: efficient speech recognition engine with compressed lstm on
  fpga.
\newblock {\em arXiv preprint arXiv:1612.00694}, 2016.

\bibitem{DBLP:journals/corr/HeZRS15}
K.~He, X.~Zhang, S.~Ren, and J.~Sun.
\newblock Deep residual learning for image recognition.
\newblock In {\em CVPR}, 2016.

\bibitem{DBLP:journals/corr/abs-1711-05189}
E.~Hesamifard, H.~Takabi, and M.~Ghasemi.
\newblock Cryptodl: Deep neural networks over encrypted data.
\newblock {\em arXiv preprint arXiv:1711.05189}, 2017.

\bibitem{marble}
J.~C. Ho, J.~Ghosh, and J.~Sun.
\newblock Marble: high-throughput phenotyping from electronic health records
  via sparse nonnegative tensor factorization.
\newblock In {\em SIGKDD}, 2014.

\bibitem{lstm}
S.~Hochreiter and J.~Schmidhuber.
\newblock Long short-term memory.
\newblock {\em Neural Comput.}, 1997.

\bibitem{DBLP:journals/corr/IandolaMAHDK16}
F.~N. Iandola, S.~Han, M.~W. Moskewicz, K.~Ashraf, W.~J. Dally, and K.~Keutzer.
\newblock Squeezenet: Alexnet-level accuracy with 50x fewer parameters and< 0.5
  mb model size.
\newblock {\em arXiv preprint arXiv:1602.07360}, 2016.

\bibitem{DBLP:journals/corr/JiangCCOTT16}
D.~Jiang, Q.~Cai, G.~Chen, H.~Jagadish, B.~C. Ooi, K.-L. Tan, and A.~K. Tung.
\newblock Cohort query processing.
\newblock {\em PVLDB}, 2016.

\bibitem{epic}
D.~Jiang, G.~Chen, B.~C. Ooi, K.-L. Tan, and S.~Wu.
\newblock epic: an extensible and scalable system for processing big data.
\newblock {\em PVLDB}, 2014.

\bibitem{2017arXiv170304730K}
P.~W. Koh and P.~Liang.
\newblock Understanding black-box predictions via influence functions.
\newblock {\em arXiv preprint arXiv:1703.04730}, 2017.

\bibitem{2017arXiv171201208K}
T.~Kraska, A.~Beutel, E.~H. Chi, J.~Dean, and N.~Polyzotis.
\newblock The case for learned index structures.
\newblock {\em arXiv preprint arXiv:1712.01208}, 2017.

\bibitem{DBLP:conf/nips/KrizhevskySH12}
A.~Krizhevsky, I.~Sutskever, and G.~E. Hinton.
\newblock Imagenet classification with deep convolutional neural networks.
\newblock In {\em NIPS}, 2012.

\bibitem{nature_dl}
Y.~LeCun, Y.~Bengio, and G.~Hinton.
\newblock Deep learning.
\newblock {\em Nature}, 2015.

\bibitem{DBLP:journals/debu/0001ZO17}
G.~Li, M.~Zhang, G.~Chen, and B.~C. Ooi.
\newblock Towards a unified graph model for supporting data management and
  usable machine learning.
\newblock {\em {IEEE} Data Eng. Bull.}, 2017.

\bibitem{DBLP:journals/corr/abs-1708-07308}
T.~Li, J.~Zhong, J.~Liu, W.~Wu, and C.~Zhang.
\newblock Ease. ml: towards multi-tenant resource sharing for machine learning
  workloads.
\newblock {\em PVLDB}, 2018.

\bibitem{2017arXiv171106373L}
Z.~Li, C.~Wang, M.~Han, Y.~Xue, W.~Wei, L.-J. Li, and F.-F. Li.
\newblock Thoracic disease identification and localization with limited
  supervision.
\newblock {\em arXiv preprint arXiv:1711.06373}, 2017.

\bibitem{survey_medical}
G.~Litjens, T.~Kooi, B.~E. Bejnordi, A.~A.~A. Setio, F.~Ciompi, M.~Ghafoorian,
  J.~A. van~der Laak, B.~van Ginneken, and C.~I. S{\'a}nchez.
\newblock A survey on deep learning in medical image analysis.
\newblock {\em Medical image analysis}, 2017.

\bibitem{2017arXiv171200559L}
C.~Liu, B.~Zoph, J.~Shlens, W.~Hua, L.-J. Li, L.~Fei-Fei, A.~Yuille, J.~Huang,
  and K.~Murphy.
\newblock Progressive neural architecture search.
\newblock {\em arXiv preprint arXiv:1712.00559}, 2017.

\bibitem{DBLP:journals/corr/abs-1207-0143}
X.~Liu, M.~Lu, B.~C. Ooi, Y.~Shen, S.~Wu, and M.~Zhang.
\newblock Cdas: a crowdsourcing data analytics system.
\newblock {\em PVLDB}, 2012.

\bibitem{DBLP:journals/corr/ShelhamerLD16}
J.~Long, E.~Shelhamer, and T.~Darrell.
\newblock Fully convolutional networks for semantic segmentation.
\newblock In {\em CVPR}, 2015.

\bibitem{model_fool}
D.~Moosavi, M.~Seyed, A.~Fawzi, and P.~Frossard.
\newblock Deepfool: a simple and accurate method to fool deep neural networks.
\newblock In {\em CVPR}, 2016.

\bibitem{exp_fool}
A.~Nguyen, J.~Yosinski, and J.~Clune.
\newblock Deep neural networks are easily fooled: High confidence predictions
  for unrecognizable images.
\newblock In {\em CVPR}, 2015.

\bibitem{tensor_train}
A.~Novikov, D.~Podoprikhin, A.~Osokin, and D.~Vetrov.
\newblock {Tensorizing Neural Networks}.
\newblock In {\em NIPS}, 2015.

\bibitem{DBLP:journals/corr/abs-1709-06416}
S.~Palkar, J.~Thomas, D.~Narayanan, A.~Shanbhag, R.~Palamuttam, H.~Pirk,
  M.~Schwarzkopf, S.~Amarasinghe, S.~Madden, and M.~Zaharia.
\newblock Weld: Rethinking the interface between data-intensive applications.
\newblock {\em arXiv preprint arXiv:1709.06416}, 2017.

\bibitem{pansurvey}
S.~J. Pan and Q.~Yang.
\newblock A survey on transfer learning.
\newblock {\em TKDE}, pages 1345--1359, 2010.

\bibitem{DBLP:journals/corr/RussakovskyDSKSMHKKBBF14}
O.~Russakovsky, J.~Deng, H.~Su, J.~Krause, S.~Satheesh, S.~Ma, Z.~Huang,
  A.~Karpathy, A.~Khosla, M.~Bernstein, et~al.
\newblock Imagenet large scale visual recognition challenge.
\newblock {\em IJCV}, 2015.

\bibitem{DBLP:journals/corr/abs-1710-09829}
S.~Sabour, N.~Frosst, and G.~E. Hinton.
\newblock Dynamic routing between capsules.
\newblock In {\em NIPS}, 2017.

\bibitem{DBLP:journals/corr/SimonyanZ14a}
K.~Simonyan and A.~Zisserman.
\newblock Very deep convolutional networks for large-scale image recognition.
\newblock {\em arXiv preprint arXiv:1409.1556}, 2014.

\bibitem{NIPS2012_4522}
J.~Snoek, H.~Larochelle, and R.~P. Adams.
\newblock Practical bayesian optimization of machine learning algorithms.
\newblock In {\em NIPS}, 2012.

\bibitem{Stoica:EECS-2017-159}
I.~Stoica, D.~Song, R.~A. Popa, D.~Patterson, M.~W. Mahoney, R.~Katz, A.~D.
  Joseph, M.~Jordan, J.~M. Hellerstein, J.~E. Gonzalez, et~al.
\newblock A berkeley view of systems challenges for ai.
\newblock {\em arXiv preprint arXiv:1712.05855}, 2017.

\bibitem{DBLP:journals/corr/SzegedyIV16}
C.~Szegedy, S.~Ioffe, V.~Vanhoucke, and A.~A. Alemi.
\newblock Inception-v4, inception-resnet and the impact of residual connections
  on learning.
\newblock In {\em AAAI}, 2017.

\bibitem{enrbm}
T.~Tran, T.~D. Nguyen, D.~Phung, and S.~Venkatesh.
\newblock Learning vector representation of medical objects via emr-driven
  nonnegative restricted boltzmann machines (enrbm).
\newblock {\em Biomedical Informatics}, 2015.

\bibitem{DBLP:journals/corr/Wang18}
S.~Wang, T.~T.~A. Dinh, Q.~Lin, Z.~Xie, M.~Zhang, Q.~Cai, G.~Chen, W.~Fu, B.~C.
  Ooi, and P.~Ruan.
\newblock Forkbase: An efficient storage engine for blockchain and forkable
  applications.
\newblock {\em arXiv preprint arXiv:1802.04949}, 2018.

\bibitem{mm15}
W.~Wang, G.~Chen, T.~T.~A. Dinh, J.~Gao, B.~C. Ooi, K.~Tan, and S.~Wang.
\newblock {SINGA:} putting deep learning in the hands of multimedia users.
\newblock In {\em ACM MM}, 2015.

\bibitem{2018arXiv180406087W}
W.~Wang, S.~Wang, J.~Gao, M.~Zhang, G.~Chen, T.~K. Ng, and B.~C. Ooi.
\newblock Rafiki: Machine learning as an analytics service system.
\newblock {\em arXiv preprint arXiv:1804.06087}, 2018.

\bibitem{Wang2017GatedSN}
W.~Wang, N.~Yang, F.~Wei, B.~Chang, and M.~Zhou.
\newblock Gated self-matching networks for reading comprehension and question
  answering.
\newblock In {\em ACL}, 2017.

\bibitem{wangsigmodrec}
W.~Wang, M.~Zhang, G.~Chen, H.~V. Jagadish, B.~C. Ooi, and K.-L. Tan.
\newblock Database meets deep learning: Challenges and opportunities.
\newblock {\em SIGMOD Rec.}, 2016.

\bibitem{2013arXiv1312.7869W}
J.~Wei, W.~Dai, A.~Kumar, X.~Zheng, Q.~Ho, and E.~P. Xing.
\newblock Consistent bounded-asynchronous parameter servers for distributed ml.
\newblock {\em arXiv preprint arXiv:1312.7869}, 2013.

\bibitem{DBLP:journals/corr/XieBFGLN14}
P.~Xie, M.~Bilenko, T.~Finley, R.~Gilad-Bachrach, K.~Lauter, and M.~Naehrig.
\newblock Crypto-nets: Neural networks over encrypted data.
\newblock {\em arXiv preprint arXiv:1412.6181}, 2014.

\bibitem{DBLP:journals/corr/abs-1708-03888}
Y.~You, I.~Gitman, and B.~Ginsburg.
\newblock Scaling sgd batch size to 32k for imagenet training.
\newblock {\em arXiv preprint arXiv:1708.03888}, 2017.

\bibitem{exp_zfnet}
M.~D. Zeiler and R.~Fergus.
\newblock Visualizing and understanding convolutional networks.
\newblock In {\em ECCV}, 2014.

\bibitem{cikm17}
K.~Zheng, W.~Wang, J.~Gao, K.~Y. Ngiam, B.~C. Ooi, and J.~W.~L. Yip.
\newblock Capturing feature-level irregularity in disease progression modeling.
\newblock In {\em CIKM}, 2017.

\bibitem{emrphenotyping}
J.~Zhou, F.~Wang, J.~Hu, and J.~Ye.
\newblock From micro to macro: data driven phenotyping by densification of
  longitudinal electronic medical records.
\newblock In {\em SIGKDD}, 2014.

\bibitem{DBLP:journals/corr/ZhouF17}
Z.-H. Zhou and J.~Feng.
\newblock Deep forest: Towards an alternative to deep neural networks.
\newblock {\em arXiv preprint arXiv:1702.08835}, 2017.

\end{thebibliography}
\end{document}